\documentclass{bmvc2k}
\usepackage{amsfonts,amssymb,amsmath,amsthm}
\usepackage{mathrsfs} 
\usepackage{subfigure}
\usepackage{algorithm, algorithmic}
\usepackage{multirow}
\usepackage{enumitem}
\usepackage{wrapfig}
\usepackage[utf8]{inputenc}

\title{Manitest: Are classifiers really invariant?}

\addauthor{Alhussein Fawzi}{alhussein.fawzi@epfl.ch}{1}
\addauthor{Pascal Frossard}{pascal.frossard@epfl.ch}{1}

\addinstitution{
 Signal Processing Laboratory (LTS4)\\
 Ecole Polytechnique Fédérale de Lausanne (EPFL) \\
 Lausanne, Switzerland
}

\runninghead{Fawzi, Frossard}{Manitest: Are classifiers really invariant?}

\newcommand{\scalprod}[2]{\left\langle #1,#2 \right\rangle}

\newcommand{\cT}{\mathcal{T}}
\newcommand{\cM}{\mathcal{M}}

\graphicspath{{./figs/}}

\newcommand{\bb}[1]{\mathbb{#1}}

\newenvironment{example}[1][Example]{\begin{trivlist}
\item[\hskip \labelsep {\bfseries #1}]}{\end{trivlist}}

\begin{document}

\maketitle

\begin{abstract}
Invariance to geometric transformations is a highly desirable property of automatic classifiers in many image recognition tasks. Nevertheless, it is unclear to which extent state-of-the-art classifiers are invariant to basic transformations such as rotations and translations. This is mainly due to the lack of general methods that properly measure such an invariance. 
In this paper, we propose a rigorous and systematic approach for quantifying the invariance to geometric transformations of any classifier. Our key idea is to cast the problem of assessing a classifier's invariance as the computation of geodesics along the manifold of transformed images. We propose the \textit{Manitest} method, built on the efficient Fast Marching algorithm to compute the invariance of classifiers. Our new method quantifies in particular the importance of data augmentation for learning invariance from data, and the increased invariance of convolutional neural networks with depth. We foresee that the proposed generic tool for measuring invariance to a large class of geometric transformations and arbitrary classifiers will have many applications for evaluating and comparing classifiers based on their invariance, and help improving the invariance of existing classifiers.
\end{abstract}

%------------------------------------------------------------------------- 
\section{Introduction}
\label{sec:intro}
Due to the huge research efforts that have been recently deployed in computer vision and machine learning, the state-of-the-art image classification systems are now reaching performances that are close to those of the human visual system in terms of accuracy on some datasets \cite{krizhevsky2012imagenet, taigman2014deepface}. Questions emerge to what differences remain between human visual system and state-of-the-art classifiers. We focus here on one key difference, namely the problem of \textit{invariance to geometric transformations}. While the human visual system is invariant to some extent to geometric transformations, it is unclear whether automatic classifiers enjoy the same invariance properties. % The goal of this paper is to provide a rigorous and systematic way of quantifying the invariance of a given classifier.
% well known that the human visual system is robust to some extent to geometric transformations, 
% while automatic classifiers can hardly recognise objects that have been transformed. 
%; that is, by moderately translating or dilating the image of a ``car'', for example, our visual system still correctly recognizes the transformed image as a ``car''. The goal of this paper is to provide a \textit{rigorous} and \textit{systematic} way of quantifying the invariance of a given classifier.
% answer to the question: what is the invariance of a classifier to certain types of geometric transformations? 
% How to quantify the invariance of a given classifier?
%By providing a systematic tool to assess the classifiers in terms of their robustness to geometric transformations, we bridge a gap towards understanding the invariance properties of different families of classifiers, and provide a general tool for comparing classifiers in terms of invariance and not only accuracy.
% One of the major differences with human eye: we need to 
The importance of invariance in classifiers has been outlined in recent works \cite{lenc15understanding, soattoC15}, and effective solutions for transformation-invariant classifications have been proposed by either adapting the classification rules with proper distance metrics \cite{simard1998transformation, fawzi2013image, vasconcelos1998multiresolution, kokiopoulou2009minimum}, or by improving the features used for classification \cite{Bengio2012,Mallat2012, Bruna2012}. To validate such new design choices and to understand how to further improve classifiers' invariance, it becomes however primordial to develop general methods to properly measure the robustness of classifiers to geometric transformations of data samples. Previous works have proposed methods to evaluate the invariance of classifiers, either by controlled changes in simple images \cite{berkes2005slow}, or by specific tests for features of popular neural network architectures \cite{goodfellow2009measuring}. These previous studies are however limited, as they are restricted to one-dimensional transformations (e.g., rotations only), to particular types of classifiers (e.g., neural networks) or to simple images (e.g., sinusoidal images), and are based on heuristically-driven quantities. Another approach for measuring invariance consists in generating datasets with transformed images, and measuring the accuracy of classifiers on these datasets \cite{larochelle2009exploring, sohn2012learning, lecun2004learning}. This is however laborious and  involves building a novel well-designed dataset to compare all classifiers on a common ground. 
% ,  These previous studies are however limited, as they are restricted to one-dimensional transformation groups, 
%Besides, these approaches of measuring invariance do not permit to have a precise understanding of the influence of specific types of geometric transformations, nor to derive general insights on the design of future  systems with improved robustness to transformations.  

% In the absence of a riogorous and systematic framework that can be applied to all classifiers equally, the notion of invariance will not be well understood. In this paper, ...
% It is well-known that the human visual system is robust to some extent to geometric transformations of the images (e.g., rotations, translations, ...); the question we ask here is whether computer vision classifiers are also robust to these transformations of the inputs. 
% We focus in this paper on the invariance of classifiers to \textit{geometric transformations} (e.g., translations, rotations, ...). Specifically, it is well known that the human visual system is robust to some extent to these types of transformations; the question  

In this paper, we propose a principled and systematic method to measure the robustness of arbitrary image classifiers to geometric transformations. In particular, we design a new framework that can be applied to any Lie group $\mathcal{T}$ and to any classifier $f$ regardless of the particular nature of the classifier. For a given image, we define the invariance measure as the minimal distance between the identity transformation and a transformation in $\mathcal{T}$ that is sufficient to change the decision of the classifier $f$ on that image. In order to define the transformation metric, our novel key idea is to represent the set of transformed versions of an image as a manifold; the transformation metric is then naturally captured by the geodesic distance on the manifold.
%  Our key idea is to represent the set of transformed versions of an image as a \textit{manifold}.
% Our new approach is inspired from recent works that have outlined the intriguingly low robustness of classifiers to adversarial noise \cite{szegedy2013intriguing}. Instead of considering adversarial noise, we focus on geometric transformations and consider a novel invariance measure, that is defined as the \textit{minimal} geometrical transformation that is sufficient to change the decision of the classifier $f$ on an image. Our key idea is to represent the set of transformed versions of an image as a \textit{manifold}. In this case, the \emph{magnitude} of a transformation is then naturally captured by the geodesic distance on the manifold.
% transformation applied to a particular image as a manifold $\Lambda$ and then to measure the \emph{magnitude} of a transformation by the corresponding geodesic distance on that image manifold. 
Hence, for a given image, our invariance measure essentially corresponds to the minimal geodesic distance on the manifold that leads to a point where the classifier's decision is changed. A global invariance measure is then derived by averaging over a sufficiently large sample set. Equipped with our generic definition of invariance, we leverage the techniques used in the analysis of manifolds of transformed visual patterns \cite{donoho2005image, wakin2005multiscale, jacques2008geometrical} and design the Manitest method built on the efficient Fast Marching algorithm \cite{kimmel1998computing,tsitsiklis1995efficient} to compute the invariance of classifiers.
%  \textbf{we could maybe have an argument in favor of FM...} 

Using Manitest, we quantitatively show the following results: (i) The invariance of convolutional neural networks and scattering transforms largely outperform SVM classifiers, (ii) Two classifiers can have a similar accuracy, but have different invariance scores, (iii) The invariance of convolutional neural networks improves with network depth, (iv) On natural images classification task, baseline convolutional networks are not invariant to slight combinations of translations, rotations, and dilations (v) Data augmentation can dramatically increase the invariance of a classifier. The latter result is particularly surprising, as an SVM with RBF kernel trained on augmented samples can outperform the invariance of convolutional neural networks (without data augmentation) on a handwritten digits dataset. Besides these results, we showcase examples illustrating the introduced invariance scores. 
%We foresee that our new invariance metric will find numerous applications for analysing classifiers' performance and eventually lead to improving the invariance of existing classifiers. 
By providing a systematic tool to assess the classifiers in terms of their robustness to geometric transformations, we bridge a gap towards understanding the invariance properties of different families of classifiers, which will hopefully lead to building new classifiers that perform closer to the human visual system. The code of Manitest is available on the project website\footnote{\url{http://sites.google.com/site/invmanitest/}}.

% \red{Mention that these are only some results, and that we can do much more with this tool, and leave that for future research, and for people to compare there classifiers with our tool?} 
% will significantly help in designing robust classifiers that 
%\red{These analyses are limited in that they either consider a one-dimensional transformation group, restricted to a particular class of classifiers, or consider synthetic images.}
%
%  \textbf{to be completed when we have all results}, and verify several facts: (i) invariance gets better with depth for deep NN, (ii) we show that two classifiers can be perfectly equal in terms of accuracy, but have different robustness properties to geometric transformations. We also discover some facts: ????
%
%- intriguing properties, wrt to transformations
%- insights on influence of adding samples for improving robustness
%- representation depth influence confirmed in the particular case of conv NN \cite{goodfellow2009measuring}
%
%
%\textbf{Make sure to thank Gabriel and Laurent in Acks}

\vspace{-5mm}

\section{Problem formulation}
\vspace{-2mm}
\subsection{Definitions}

We consider a mathematical model where images are represented as functions $I: \bb{R}^2 \rightarrow \bb{R}$, and we denote by $L^2$ the space of square integrable images. Let $\mathcal{T}$ be a Lie group consisting of geometric transformations on $\mathbb{R}^2$, and we denote by $p$ the dimension of $\mathcal{T}$ (i.e., number of free parameters). % \footnote{In principle, our framework works for more general class of deformations, as long as these can be modeled using finite-dimensional parameters.}
For any transformation $\tau$ that belongs to $\mathcal{T}$, we denote by $I_{\tau}$ the image $I$ transformed by $\tau$. That is, $I_{\tau} (x,y) = I (\tau^{-1} (x,y))$.
% ; for any $\tau \in \mathcal{T}$, we denote by $I_{\tau}$ the image $I$ transformed by $\tau$. We assume that $\mathcal{T}$ is embedded in $\bb{R}^p$; hence any $\tau \in \mathcal{T}$ can be represented by a $p$-dimensional vector. 
Examples of Lie groups include the rotation group SO(2) ($p=1$, described by one angle) and the similarity group ($p=4$, described by a 2D translation vector, a dilation and an angle). 

% We define $d: \mathcal{T} \times \mathcal{T} \rightarrow \mathbb{R}^+$ to be an arbitrary metric on $\mathcal{T}$, and denote by $e$ the identity element of the group $\mathcal{T}$. 
% We refer to $d(e, \tau)$ as the \textit{magnitude} of a transformation $\tau$, where $d: \mathcal{T} \times \mathcal{T} \rightarrow \bb{R}^+$ is an arbitrary metric on $\mathcal{T}$.

  % In other words,  given a pair of transformations $\tau_1, \tau_2$, the \textit{distance} between the two transformations is denoted by $d(\tau_1, \tau_2)$.

% In the following, we will need to define distances on $\mathcal{T}$.
% $d: \mathcal{T} \times \mathcal{T} \rightarrow \bb{R}^+$  arbitrary metric that measures the distance between two transformations in $\mathcal{T}$.
% We consider a classification task with $L$ labels, and $f$ ...

Consider an image classification task, where the images are assigned discrete labels in $\mathcal{L} = \{1, \dots, L\}$, and let $f$ be an arbitrary image classifier. Formally, $f$ is a function defined on the space of square integrable images $L^2$, and takes values in the set $\mathcal{L}$. 
Our goal in this paper is to evaluate the invariance of $f$ with respect to $\mathcal{T}$. Given an image $I$, we define the invariance score of $f$ relative to $I$, $\Delta_{\mathcal{T}} (I; f)$, to be the \textit{minimal normalized distance} from the identity transformation to a transformation $\tau$ that changes the classification label, i.e., 
\begin{align}
\label{eq:delta_f}
\Delta_{\mathcal{T}} (I; f) = \min_{\tau \in \mathcal{T}} \frac{d(e, \tau)}{\| I \|_{L^2}} \text{ subject to } f(I_{\tau}) \neq f(I),
\end{align}
where $e$ is the identity element of the group $\mathcal{T}$ and $d: \mathcal{T} \times \mathcal{T} \rightarrow \bb{R}^+$ is a metric on $\mathcal{T}$ that we define later (Section \ref{sec:transformation_metric}).
% \red{Justify min + justify normalization}
% \red{Comment on normalization?}
The invariance score quantifies the resilience of $f$ to transformations in $\mathcal{T}$, namely larger values of $\Delta_{\mathcal{T}} (I; f)$ indicate a larger invariance.
It is worth noting that our definition of $\Delta_{\mathcal{T}}$ is related to the recent work in \cite{szegedy2013intriguing} that defined adversarial noise as the minimal perturbation (in the Euclidean sense) required to misclassify the datapoint. However, instead of considering generic adversarial perturbations, we focus on minimal \textit{geometric transformations}, with a metric borrowed from the group $\mathcal{T}$.
% outlined the the intriguingly low robustness of classifiers to adversarial noise. However, instead of considering adversarial noise considering adversarial noise, we focus on geometric transformations 
% is defined as the minimal perturbation (in the $\ell_2$ sense) that switches the estimated lab el.

For a given a distribution of datapoints $\mu$, the global invariance score of $f$ to transformations in $\mathcal{T}$ is defined by
\begin{align}
\rho_{\mathcal{T}} (f) = \bb{E}_{I \sim \mu} \Delta_{\mathcal{T}} (I;f).
\end{align}
% By denoting $\mu$ the distribution of images in the considered task, a global measure to the resilience of $f$ to transformations in $\mathcal{T}$ is defined as follows: 
% The normalization with the $2$-norm has the role of equalizing the contributions of different images, independently of their norm.
The quantity $\rho_{\mathcal{T}} (f)$ depends on $f$ as well as the distribution of datapoints $\mu$. However, to simplify notations, we have omitted the dependence on $\mu$, assuming the distribution is clear from the context. In practical classification tasks, the true underlying distribution $\mu$ is generally unknown. In that case, we estimate the global resilience by taking the empirical average\footnote{In practice, it is sufficient to consider an empirical average over a sufficiently large random subset of the training set. The number of samples is chosen to achieve a small enough confidence interval.}
over training points: $\hat{\rho}_{\mathcal{T}} (f) = \frac{1}{m} \sum_{j=1}^m \Delta_{\mathcal{T}} (I_j;f)$.
% reaches the true with high confidence % consider instead the empirical average $\mu$ is not known in practice, we consider instead an empirical average over all training points 
% \begin{align}
% \end{align}
\vspace{-3mm}
\subsection{Transformation metric}
\label{sec:transformation_metric}
% \red{For Figure 1, I can generate the set of images that go along the geodesic curve between $I_{\tau_0}$ and $I_{\tau_3}$ (i.e., the mobile that gets slowly translated and rotated from $I_{\tau_0}$ till it becomes $I_{\tau_3}$ (where a small local transformation is applied between two consecutive images). Do you think this can help understanding/motivating the notions of geodesic?}

\begin{figure}[t]
\centering
\includegraphics[width=0.5\textwidth]{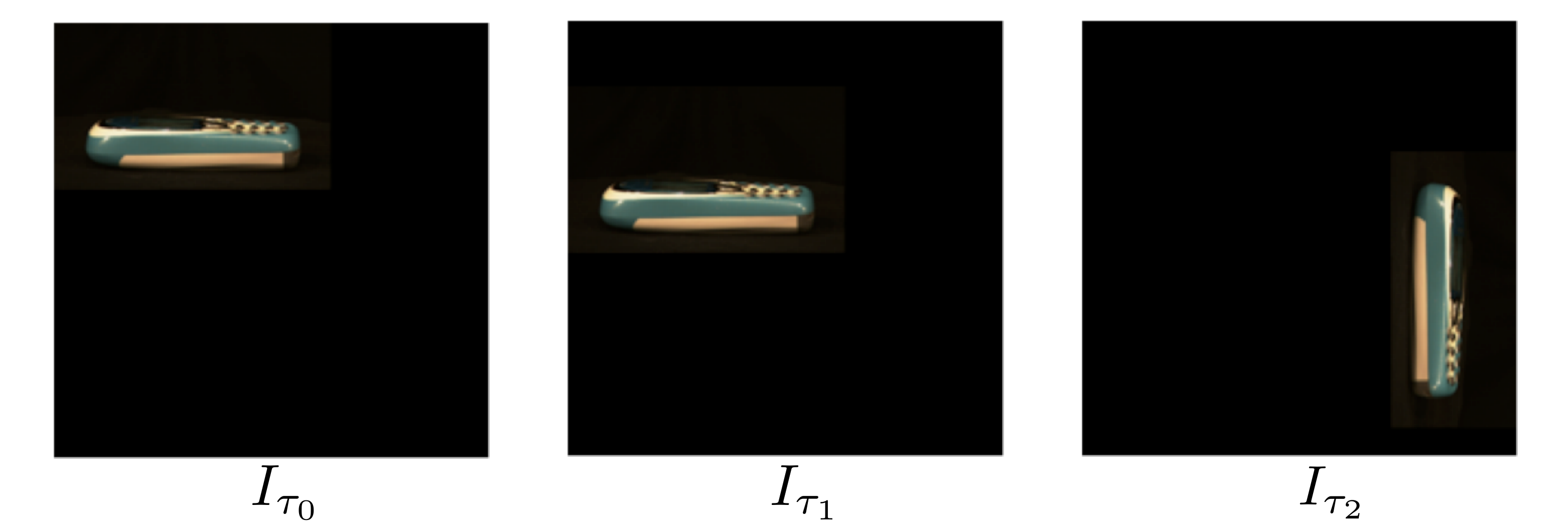}
\caption{\label{fig:fig_L2}Schematic representation of the problem encountered by using metric the $L^2$ metric. Black pixels indicate pixels with value $0$, and $I_{\tau_1}, I_{\tau_2}$ are obtained by applying a combination of rotation and translation to $I_{\tau_0}$.  Image taken from \cite{geusebroek2005amsterdam}.} 
% Using $d_I$, note that $\tau_{1}$ and $\tau_{2}$ are equidistant to $\tau_0$. However, we require to have $d(\tau_0, \tau_1) < d(\tau_0, \tau_2)$, as $I_{\tau_2}$ clearly underwent a more severe transformation than $I_{\tau_1}$.}
\end{figure}

\begin{figure}[t]
\centering
\includegraphics[width=1\textwidth]{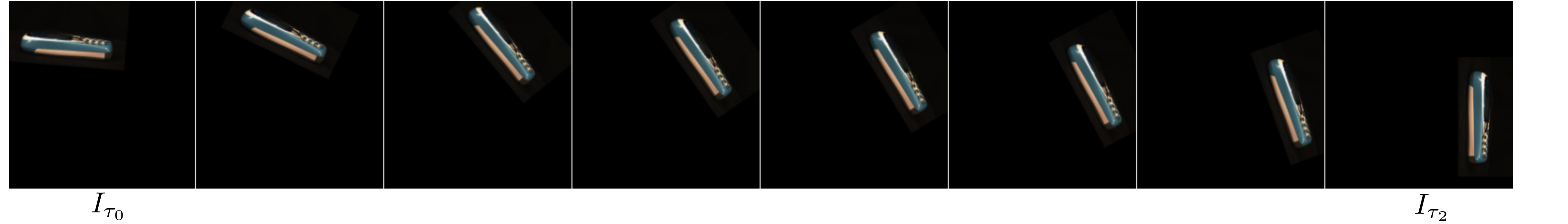}
\caption{\label{fig:fig_mobile_trajectory} Images along the geodesic path from $I_{\tau_0}$ to $I_{\tau_2}$}
\end{figure}

% For a given image $I$, the resilience of $f$ to geometric transformations $\Delta_{\mathcal{T}} (I; f)$ heavily depends on the choice of the metric $d$ that we impose on $\mathcal{T}$. % We formally define the distance that we use in the paper. 
We discuss and introduce the distance used for the invariance score $\Delta_{\mathcal{T}} (I; f)$. 
It should be noted that $\mathcal{T}$ is possibly a multi-dimensional group (i.e., the transformations in $\mathcal{T}$ are described by many parameters of different nature such as translation, rotation, scale, ...); hence, defining a trivial metric that measures the absolute distance between transformation parameters is of limited interest, as it combines parameters possibly of different nature.
 Instead, a more relevant notion of distance is one that \textit{depends on the underlying image} $I$. In that case, $d(\tau_1, \tau_2)$ quantifies the change in \textit{appearance} between images $I_{\tau_1}$ and $I_{\tau_2}$, rather than an absolute distance between the two transformations. Consider for example the \textit{image distance} $d_{I}(\tau_1, \tau_2) = \| I_{\tau_1} - I_{\tau_2} \|_{L^2}$. %\footnote{Large enough sample. We compute this average on the sample of the data to achieve a small enough confidence interval.} 
%\begin{align}
%\label{eq:metric_dI}
%d_{I}(\tau_1, \tau_2) = \| I_{\tau_1} - I_{\tau_2} \|_{L^2}.
%\end{align}
While $d_I$ explicitly depends on the underlying image $I$, it fails to capture the intrinsic geometry of the family of transformed images. To illustrate this point, we consider a simple example of images in Fig. \ref{fig:fig_L2} with two transformed versions $I_{\tau_1}$ and $I_{\tau_2}$ of a reference image $I_{\tau_0}$. Note that $d_I(\tau_0, \tau_1) = d_I(\tau_0, \tau_2)$, as both transformed objects have no intersection with the reference object. % as the transformed images are completely distinct from the pixels of the reference image.
% While $d_I(\tau_0, \tau_1) = d_I(\tau_0, \tau_2)$, 
However, it is clear that $I_{\tau_2}$ incurred a large rotation and translation, while $I_{\tau_1}$ underwent a slight vertical translation. Hence, the distance metric should naturally satisfy $d(\tau_0, \tau_1) < d(\tau_0, \tau_{2})$, which is not the case for the image distance. This is crucial in our setting, as a classifier that recognizes the similarity of the objects in $I_{\tau_2}$ and $I_{\tau_0}$ is certainly more robust to transformations than a classifier that merely recognizes the similarity between $I_{\tau_1}$ and $I_{\tau_0}$, and should be given a higher score.
% This property of the metric is all the more important in our settings, as a classifier that recognizes the similarity of the objects in $I_{\tau_2}$ and $I_{\tau_0}$ is certainly more robust to rotations and translations than a classifier that merely recognizes the similarity between $I_{\tau_1}$ and $I_{\tau_0}$.
% \footnote{This property of the metric is important in our settings, as a classifier that recognizes the similarity between images $I_{\tau_3}$ and $I_{\tau_0}$ should be given a larger score than one that ``only'' sees the similarity between $I_{\tau_0}$ and $I_{\tau_1}$.}.
% , as the translation applied to $I_{\tau_1}$ is much smaller than the one applied to $I_{\tau_2}$ or $I_{\tau_3}$. 
This example underlines a well-known fundamental issue with the $L^2$ distance that fails to capture the intrinsic distance of the curved manifold of transformed images (see e.g., \cite{tenenbaum2000global, donoho2005image}). To correctly capture the intrinsic structure of the manifold, we define $d$ to be the length of the shortest path belonging to the manifold (i.e., the \textit{geodesic distance}).  For illustration, we show in Fig. \ref{fig:fig_mobile_trajectory} images along the geodesic path from $\tau_0$ to $\tau_2$; the geodesic distance is then essentially the sum of \textit{local} $L^2$ distances between transformed images over the geodesic path. We formalize these notions as follows.
% in that it obscures the intrinsic distance of the manifold of transformed images. In fact, the family of transformed images represents a \textit{curved subset} of $L^2$ (called Image Appearance Manifold), and the intrinsic geometry of this manifold is obtained through the geodesic distance. In the following, we formalize this.
% Fig. \ref{}, where an image $I_{\tau_0}$ 
% the $L^2$ distance is well
% \red{Consider for example the following distance:} $L^2$. Actually, we want the smallest curve on a manifold called the appearance manifold. In the following, we formalize this:
% Informally, with such a definition, the weight given to rotating a nearly radial object will be much smaller than that of translating it, as the latter has a larger effect on the appearance of the object. 

Let $\mathcal{M} (I)$ be the family of transformed images $\mathcal{M} (I) = \{ I_{\tau}: \tau \in \mathcal{T} \}$.
%  \textit{Image Appearance Manifold (IAM)} to be the family of images obtained by applying transformations in $\mathcal{T}$ to $I$:
% \begin{align*}
% \mathcal{M} (I) = \{ I_{\tau}: \tau \in \mathcal{T} \}.
% \end{align*}
Equipped with the $L^2$ metric, $\mathcal{M} (I)$ defines a metric space and a continuous submanifold of $L^2$. Following the works of \cite{wakin2005multiscale, jacques2008geometrical} that considered similar manifolds in different contexts, we call $\mathcal{M} (I)$ an \textit{Image Appearance Manifold} (IAM), and we follow here their approach.
Assuming that $\gamma: [0,1] \mapsto \mathcal{T}$ is a $C^1$ curve in $\mathcal{T}$, and that $I_{\gamma(t)}$ is differentiable with respect to $t$, we define the \textit{length} $L(\gamma)$ of $\gamma$ as
\vspace{-2mm}
\begin{align}
\label{eq:Lgamma}
L(\gamma) = \int_{0}^1 \left\| \frac{d}{dt} I_{\gamma(t)} \right\|_{L^2} dt. \end{align}
Note that Eq. (\ref{eq:Lgamma}) is expressed in terms of the $L^2$ metric in the image appearance manifold and corresponds to summing the local $L^2$ distances between transformed images over the path $I_{\gamma}$. We now show that $L(\gamma)$ can be expressed as a length associated to a Riemannian metric on $\mathcal{T}$ that we now derive.
% sum of local L^2 distances between the transformed images
% We show that the length of the curve $t \rightarrow I_{\gamma(t)}$ in $\mathcal{M}(I)$ is equal to  the length of the curve $t \rightarrow \gamma(t)$ on $\mathcal{T}$, for an appropriate Riemannian metric that we now derive.
% In other words, $L(\gamma)$ denotes the length of the curve $t \rightarrow I_{\gamma(t)}$ in $\mathcal{M}(I)$. 
Defining the map \[ F:\cT \rightarrow \cM, \quad \tau \mapsto I_{\tau}, \]we have
\[
\frac{d}{dt} I_{\gamma(t)} = (F\circ \gamma)'(t) = dF_{\gamma(t)}(\gamma'(t)),
\]
where $dF_{\tau}$ denotes the differential of $F$ at $\tau$, and $\gamma'$ is derivative of $\gamma$. It follows that
\[
L(\gamma) = \int_{0}^{1} \sqrt{g_{\gamma(t)}(\gamma'(t),\gamma'(t))} dt
\]
where $g_{\tau}$ is the \textit{Riemannian metric} (i.e., a positive bilinear form on $T_{\tau}\mathcal{T}$, the tangent space of $\mathcal{T}$ at $\tau$), given by:
\[
g_{\tau}(v,w) = \scalprod{dF_{\tau}(v)}{dF_{\tau} (w)}_{L^2} \text{ for all } v,w \in T_\tau\mathcal{T}.
\]
Note that $g$ can be equivalently seen as the pullback of the $L^2$ metric on $\mathcal{M} (I)$ along $F$.
% The Riemannian manifold $(\mathcal{T}, g)$ can be equivalently seen as an immersed manifold \cite{do1992riemannian} in the image appearance manifold $\mathcal{M} (I)$ equipped with the $L^2$ metric. The metric $g$ is then the associated pullback metric. % \red{Put in footnote maybe that} 
By choosing a basis in the tangent space, the length $L(\gamma)$ can be equivalently written
\[
L(\gamma) = \int_{0}^{1} \sqrt{\gamma'(t)^T G_{\gamma(t)} \gamma'(t)} dt,
\]
where $G_{\gamma(t)}$ is the $p \times p$ positive definite matrix associated to the bilinear form $g$. % \red{Instead of putting $\gamma'$ above, put the representation of $\gamma'$ in the chosen basis?}.

\begin{example}[Example 1 (Rotation, $\mathcal{T} = SO(2)$)] The transformation group $\mathcal{T}$ is parametrized with a rotation angle $\theta$ ($p=1$). In this case, the matix $G_{\theta}$ is of size $1$ by $1$, and equal to
\[
G_{\theta} = \left\| \frac{\partial I_{\theta}}{\partial \theta} \right\|^2_{L^2}. \qed
\]
\end{example}

% Let $\tau_h$ be the rotation of angle $h$. Then, we have $dF_{\tau}(v) = \frac{I_{\tau_h} - I}{h}$. (v?).

\begin{example}[Example 2 (Dilation+Rotation).]
The group $\mathcal{T}$ has $2$ degrees of freedom; namely a scale parameter $a$, and a rotation angle $\theta$. The Riemannian metric reads
\[
G_{\tau} = 
\begin{bmatrix} 
\scalprod{\frac{\partial I_{\tau}}{\partial a}}{\frac{\partial I_{\tau}}{\partial a}} & \scalprod{\frac{\partial I_{\tau}}{\partial a}}{\frac{\partial I_{\tau}}{\partial \theta}} \\ \scalprod{\frac{\partial I_{\tau}}{\partial \theta}}{\frac{\partial I_{\tau}}{\partial a}} & \scalprod{\frac{\partial I_{\tau}}{\partial \theta}}{\frac{\partial I_{\tau}}{\partial \theta}}
\end{bmatrix}. \qed
\]
%\[
%G_{\tau} = 
%\begin{bmatrix} 
%\scalprod{\frac{\partial I_{\tau}}{\partial t_x}}{\frac{\partial I_{\tau}}{\partial t_x}} & \scalprod{\frac{\partial I_{\tau}}{\partial t_x}}{\frac{\partial I_{\tau}}{\partial t_y}} & \scalprod{\frac{\partial I_{\tau}}{\partial t_x}}{\frac{\partial I_{\tau}}{\partial a}} & \scalprod{\frac{\partial I_{\tau}}{\partial t_x}}{\frac{\partial I_{\tau}}{\partial \theta}} \\
%\scalprod{\frac{\partial I_{\tau}}{\partial t_y}}{\frac{\partial I_{\tau}}{\partial t_x}} & \scalprod{\frac{\partial I_{\tau}}{\partial t_y}}{\frac{\partial I_{\tau}}{\partial t_x}} & \scalprod{\frac{\partial I_{\tau}}{\partial t_y}}{\frac{\partial I_{\tau}}{\partial a}} & \scalprod{\frac{\partial I_{\tau}}{\partial t_y}}{\frac{\partial I_{\tau}}{\partial \theta}} \\
%\scalprod{\frac{\partial I_{\tau}}{\partial a}}{\frac{\partial I_{\tau}}{\partial t_x}} & \scalprod{\frac{\partial I_{\tau}}{\partial a}}{\frac{\partial I_{\tau}}{\partial t_y}} & \scalprod{\frac{\partial I_{\tau}}{\partial a}}{\frac{\partial I_{\tau}}{\partial a}} & \scalprod{\frac{\partial I_{\tau}}{\partial a}}{\frac{\partial I_{\tau}}{\partial \theta}} \\
%\scalprod{\frac{\partial I_{\tau}}{\partial \theta}}{\frac{\partial I_{\tau}}{\partial t_x}} & \scalprod{\frac{\partial I_{\tau}}{\partial \theta}}{\frac{\partial I_{\tau}}{\partial t_y}} & \scalprod{\frac{\partial I_{\tau}}{\partial \theta}}{\frac{\partial I_{\tau}}{\partial a}} & \scalprod{\frac{\partial I_{\tau}}{\partial \theta}}{\frac{\partial I_{\tau}}{\partial \theta}}
%\end{bmatrix}. \qed
%\]
\end{example}

Having defined the length of a curve on $\mathcal{T}$, the geodesic distance between two points $\tau_1, \tau_2$ is defined as the length of the shortest curve joining the two points:
\[
d(\tau_1, \tau_2) = \inf \{L(\gamma): \gamma \in C^1([0,1]), \gamma(0) = \tau_1, \gamma(1) = \tau_2\}.
\]

Finally, our problem therefore consists in computing the global invariance score, or equivalently $\Delta_{\mathcal{T}} (I; f)$ defined in Eq. (\ref{eq:delta_f}), where $d$ is the geodesic distance. In other words, our problem becomes that of computing the minimal geodesic distance from the identity transformation to a transformation that is sufficient to change the estimated label of $f$.

\vspace{-4mm}
% \section{Invariance score computation}
\section{Invariance score computation}
% For a positive semi-definite matrix $A$, we define $\| x \|_A^2 = x^T A x$, and $\scalprod{x}{y}_{A} = x^T A y$ the inner produce associated to the tensor $A$.
% In this section, we describe the proposed algorithm to compute the invariance score $\Delta_{\mathcal{T}} (I; f)$, where $d$ is the geodesic distance previously defined. The main difficulty of optimizing over $\tau$ in the expression of $\Delta_{\mathcal{T}} (I; f)$ lies in the computation of the geodesic distance. Moreover, it should be noted that the constraints are highly nonlinear, as the composition operator of $\mathcal{T}$, as well as the classifier $f$ are involved. 
The key to an efficient and accurate approximation of $\Delta_{\mathcal{T}} (I; f)$ lies in the effective computation of geodesics on the manifold $(\mathcal{T}, G)$ that we address as follows. % In this section, we review the FM method, and apply it to our setting for the computation of robustness to transformations.
% We first review the Fast Marching algorithm for the computation of geodesics on manifolds (Sec. \ref{sec:fast_marching}), and apply it to our setting in Sec. \ref{sec:proposed_algorithm}

% The optimization problem defined by $\Delta_{\mathcal{T}} (I; f)$ is challenging as the objective function involves a geodesic computation, and the constraint is nonlinear. For a given Riemannian manifold $\mathcal{T}$ equipped with a metric $G$, 
% \subsection{Fast Marching for computation of geodesics on manifolds}
% \label{sec:fast_marching}

Let $u(\tau) = d(e, \tau)$ be the \textit{geodesic map} that measures the geodesic distance between the (fixed) identity element and $\tau$. The geodesic map satisfies the following Eikonal equation \cite{peyre2010geodesic}
\begin{align}
\label{eq:eikonal_equation}
\| \nabla u (\tau) \|_{G_\tau^{-1}} = 1 \text{ for } \tau \in \mathcal{T}\backslash\{e\}, \text{ and } u(e) = 0,
\end{align}
where $\| x \|_A = \sqrt{\scalprod{x}{x}_A}$ with $\scalprod{x}{y}_A = x^T A y$. Moreover, it was proved in \cite{crandall1983viscosity} that the geodesic map $u$ is the \textit{unique} viscosity solution of the Eikonal equation, provided that $\tau \rightarrow G(\tau)$ is continuous. Many numerical schemes rely on the Eikonal equation characterization to approximate the geodesic map. We use here the popular \textit{Fast Marching (FM) method} \cite{kimmel1998computing}, a fast front propagation approach that computes the values of the discrete geodesic map in increasing order. % We only provide here a brief description of FM due to space constraints, and we refer to the survey in \cite{peyre2010geodesic} for a more complete explanation.
We only provide here a brief description of FM due to space constraints, and focus on the case where the manifold $\mathcal{T}$ is two-dimensional (i.e., $p=2$). The extension to arbitrary dimensions is straightforward, and we refer to \cite{peyre2010geodesic, sethian2003ordered} for more complete explanations and computations.
% the Eikonal equation characterization to approximate the geodesic map. 
% on solving a discrete version of the Eikonal equation to approximate the geodesic map. 
 \begin{wrapfigure}{r}{0.4\textwidth}
  \begin{center}
    \includegraphics[width=0.3\textwidth]{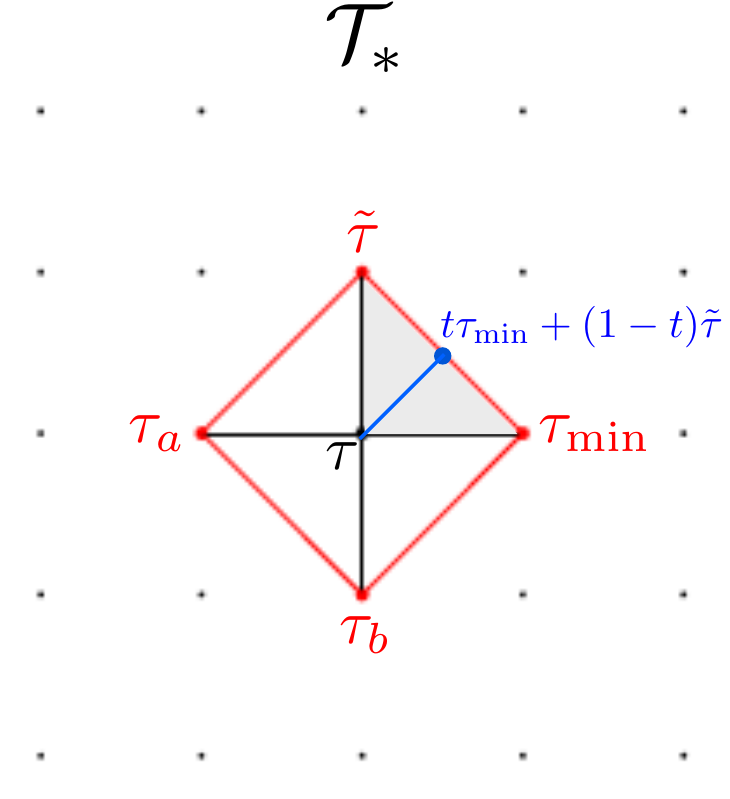}
  \end{center}
  \caption{\label{fig:stencil} Schematic representation of the discretized manifold $\mathcal{T}_*$, and the Fast Marching update rule. In this figure, we have $\mathcal{N} (\tau) = \{ \tilde{\tau}, \tau_{\min}, \tau_a, \tau_b \}$.}
\end{wrapfigure}We assume that the manifold $\mathcal{T}$ is sampled using a regular grid; let $\mathcal{T}_{*}$ be the sampling of $\mathcal{T}$, and $U$ be the discrete vector that approximates $u$ at the nodes. The structure of Fast Marching is almost identical to Dijkstra's algorithm for computing shortest paths on graphs \cite{dijkstra1959note}. The main difference lies in the update step, which bypasses the constraint of propagation along edges. For a given node $\tau$, define $\mathcal{N}(\tau)$ to be the set of neighbours of $\tau$ (see illustration in Fig. \ref{fig:stencil}). 
%Moreover, define the set of reversed neighbours $\mathcal{N}[\tau]$
%\begin{align*}
%\mathcal{N} [\tau] = \{ \tau'' \in \mathcal{T}_{*}: \exists \tau' \in \mathcal{T}_* \text{ such that } (\tau, \tau') \in \mathcal{N} (\tau'') \}.
%\end{align*}
In the FM algorithm, each grid point is tagged either as \textit{Known} (nodes for which distance is frozen), or \textit{Unknown} (nodes for which distance can change in subsequent iterations). Initially, the grid points are set to \textit{Unknown}, and $U$ is set to $\infty$, except $U(e)$ that is set to zero. At each iteration of FM, the unknown node $\tau_{\text{min}}$ with smallest $U$ is selected, and tagged as \textit{Known}. Then, each unknown neighbour $\tau \in \mathcal{N} (\tau_{\min})$ is visited, and $U(\tau)$ is updated as follows: $U(\tau)$ is set to be the minimum of itself, $U(\tau_{\min}) + \| \tau - \tau_{\min} \|_{G_\tau}$ and 
\begin{align}
\label{eq:update_FM}
\min_{t \in [0,1]} t U(\tau_{\min}) + (1-t) U(\tilde{\tau}) + \|t \tau_{\text{min}} + (1-t) \tilde{\tau} - \tau \|_{G_{\tau}},
\end{align}
for each known $\tilde{\tau}$ such that $(\tau, \tau_{\min}, \tilde{\tau})$ forms a triangle (see Fig. \ref{fig:stencil}).
% for each $\tilde{\tau}$ such that $(\tau_{\min}, \tilde{\tau}) \in \mathcal{N}(\tau)$. 
It is worth noting that, unlike Dijkstra, FM seeks the optimal point (possibly outside the set $\mathcal{T}_*$) on the neighbourhood boundary that minimizes the estimated distance at $\tau$, under a linear approximation assumption (Eq. \ref{eq:update_FM}). Fortunately, the problem in Eq. (\ref{eq:update_FM}) can be solved in closed form, as it corresponds to the minimization of a scalar quadratic equation \cite{sethian2003ordered}.

\begin{algorithm}[t]
\caption{Manitest method (with $p=2$) for computing $\Delta_{\mathcal{T}} (I; f)$}
\label{alg:algo_fm_invariance}
\begin{algorithmic}
\STATE  Initialize $U(e) = 0$, $U = \infty$ otherwise, and tag all nodes as \textit{unknown}.
\WHILE{termination criterion is not met}
\STATE Select the \textit{unknown} node $\tau_{\min}$ that achieves minimal distance $U$.
\STATE Tag $\tau_{\min}$ as \textit{known}.
\STATE If $f(I_{\tau_{\min}}) \neq f(I)$, set $\Delta_\mathcal{T}(I;f) \leftarrow U(\tau_{\min}) / \| I \|_{L^2}$ and terminate.
\FORALL{\textit{unknown} $\tau \in \mathcal{N} (\tau_{\min})$}
\STATE Update $U(\tau)$ to be the minimum of itself, $U(\tau_{\min}) + \| \tau - \tau_{\min} \|_{G_\tau}$ and the expression in Eq.(\ref{eq:update_FM}).
%\begin{align*}
%\end{align*}
\ENDFOR
\ENDWHILE
\end{algorithmic}
\end{algorithm}

% If \textit{Trial} points are ordered in a min-heap data structure, the computational complexity of FM is $O(N \log(N))$, where $N$ is the number of points on the grid.

% \noindent \textbf{Remarks about Algorithm \ref{alg:algo_fm_invariance}:}

% \textit{Stopping criterion:}  
The Manitest method, which applies FM algorithm to compute $\Delta_{\mathcal{T}} (I; f)$, is given in Algorithm \ref{alg:algo_fm_invariance} in the two dimensional case. The algorithm is stopped whenever a transformation that changes the classification label is found.%; this is possible since FM visits the nodes in an increasing order of $U$.
\footnote{To ensure the termination of the algorithm (even if no successful transformation is found) we limit the number of iterations $N$ to $50,000$. However, in all our experiments, this limit was never reached, and the algorithm terminated by successfully finding a transformation that satisfies $f(I_{\tau}) \neq f(I)$.} The nodes and metrics are generated on-the-fly in order to avoid spending unnecessary ressources on far-away nodes that might be farther than the minimal transformation that satisfies $f(I) \neq f(I_\tau)$ and therefore never visited. 

The complexity of Manitest is $O(N\log(N))$, where $N$ is the number of visited nodes if a min-heap structure is used \cite{peyre2010geodesic} (for constant $p$, and constant cost for evaluation of $f$). It is important to note however that the complexity of the algorithm has an exponential dependence on the dimension $p$ since our method involves the enumeration of simplices in dimension $p$; this is however not a big limitation as our main focus goes to low-dimensional transformation groups (e.g., $p \leq 6$ for affine transformations).

%Note that the number of simplices depend exponentially on the dimension $p$; this is however not a
%Due to the exponential dependence on $p$, our approach is therefore adapted to low-dimensional transformation groups (e.g., 2D similarity or affine transformations), which can include nonetheless a rich set of transformations.

% \textit{Complexity:} If a min-heap structure is used, the complexity of Algorithm \ref{alg:algo_fm_invariance} is $O(N(\log(N) + p3^p))$, where $N$ is the number of visited nodes \cite{peyre2010geodesic}. Due to the exponential dependence on $p$, our approach is therefore adapted to low-dimensional transformation groups (e.g., 2D similarity or affine transformations), that can include nonetheless a rich set of transformations.
% \textit{Consistency:} 
Finally, we note that when the metric is isotropic (i.e., $G_{\tau}$ is proportional to the identity matrix for all $\tau$), FM provides a consistent scheme.  That is, as the discretization step tends to zero, the solution computed by the algorithm tends towards the viscosity solution of the Eikonal equation. Unfortunately, for arbitrary anisotropic metrics, consistency is however not guaranteed, and the exact computation of the geodesics becomes much more difficult and computationally demanding (see \cite{sethian2000fast, benmansour2011tubular, lin2003enhancement, mirebeau2014anisotropic}). 
% Several improvements over classical FM were proposed to deal with anisotropy. However, these methods usually extend the neighbourhood, and potentially make the scheme inefficient. 
% In our work, despite the anisotropy of $G$, we use the classical FM algorithm on a regular grid. 
However, we observed that the anisotropy of the considered metric is generally not very large in the vicinity of $e$ (although it exceeds the theoretical limit of guaranteed consistency). This leads to empirically accurate estimates of the geodesic distance using Manitest, when the discretization step is sufficiently small. Finally, we stress that that all previous methods addressing the metric anisotropy can readily be applied to our setting, and we leave that as future work.

\vspace{-4mm}
\section{Experiments}
% \red{In intro, say that hostile environments, need to evaluate it when applying different transformations. Needs to be robust}
% In this section, we assess and compare the robustness of different classifiers using the proposed approach.
% \red{should we do a settings paragraph? At least, I should be saying somewhere how I sampled that transformation manifolds (i.e., rotation with a step size of $\pi/20$, translations with half a pixel, and dilation with $0.1$ discretization.}
% \red{Also, say somewhere that I am padding with zeros}

We propose now a set of experiments to study the invariance of classifiers in different settings. In particular, we consider the following transformation groups: 
\begin{itemize}[noitemsep,nolistsep]
\item $\mathcal{T}_{\text{trans}}$: in-plane translations of the image ($p=2$),
\item $\mathcal{T}_{\text{dil+rot}}$: dilations and rotations around the center of the image ($p=2$),
\item $\mathcal{T}_{\text{sim}}$: similarity transformations that describe combinations of translations, dilations and rotations around the center of the image ($p=4$). 
\end{itemize}

In all experiments, we used a discretization step of $0.5$ pixels for translations, $\pi/20$ radians for rotation, and $0.1$ for dilation for Manitest. % These parameters chosen to achieve precise enough geodesic estimation, while remaining computationally efficient. 
Finally, the transformed images have the same size as the original image, and we use a zero-padding boundary condition.

\vspace{-3mm}
\subsection{Handwritten digits dataset}
\label{sec:mnist}
We first compare the invariance of different classifiers on the MNIST handwritten digits dataset \cite{lecun1998gradient}.
We consider the following classifiers:
\begin{enumerate}[noitemsep,nolistsep]
\item \textbf{Linear SVM} \cite{fan2008liblinear},
\item \textbf{SVM with RBF kernel} \cite{chang2011libsvm},
\item \textbf{Convolutional Neural Network} \cite{vedaldi2014matconvnet}: we employ a baseline architecture with two hidden layers containing each a convolution operation ($5 \times 5$ filters with 32 feature maps for the first layer and $64$ for the second layer), a rectified linear unit nonlinearity, and a max pooling over $2 \times 2$ windows followed by a subsampling. The architecture is trained with stochastic gradient descent, with a softmax loss.
\item \textbf{Scattering transform followed by a generative PCA classifier}. We used the same settings as in \cite{bruna2013invariant}, and we refer to that paper for more details. %  we used the same classifier as (CITE) for digits classification, and we refer to it for more details.
\end{enumerate}

%\begin{small}
%\begin{table}
%\begin{center}
%\begin{tabular}{|l|c|c|c|c|}
%\hline
%Group & L-SVM & RBF-SVM & CNN & Scat. PCA \\
%\hline\hline
%% Train error $(\%)$ & 7.7 & 0 & ... & ... \\ \hline
%Test error $(\%)$ & 8.4 & 1.4 & \textbf{0.7} & 0.8 \\ \hline\hline
%Translations ($\mathcal{T} = \mathcal{T}_{\text{trans}}$) & 0.76 & 1.23 & 1.70 & \textbf{2.0} \\ \hline
%Dilations + Rotations ($\mathcal{T} = \mathcal{T}_{\text{dil+rot}}$) & 0.81 & 1.48 & \textbf{1.84} & 1.77 \\ \hline
%Similarity ($\mathcal{T} = \mathcal{T}_{\text{sim}}$) & 0.60 & 1.05 & 1.33 & \textbf{1.50}\\ \hline
%\end{tabular}
%\end{center}
%\caption{\label{tab:mnist_results} Accuracy and invariance scores of different classifers on the MNIST dataset.}
%\end{table}
%\end{small}

% AFTER CORRECTION OF CODE
\begin{small}
\begin{table}
\begin{center}
\begin{tabular}{|l|c|c|c|c|}
\hline
Group & L-SVM & RBF-SVM & CNN & Scat. PCA \\
\hline\hline
% Train error $(\%)$ & 7.7 & 0 & ... & ... \\ \hline
Test error $(\%)$ & 8.4 & 1.4 & \textbf{0.7} & 0.8 \\ \hline\hline
Translations ($\mathcal{T} = \mathcal{T}_{\text{trans}}$) & 0.8 & 1.3 & 1.7 & \textbf{2.1} \\ \hline
Dilations + Rotations ($\mathcal{T} = \mathcal{T}_{\text{dil+rot}}$) & 0.8 & 1.5 & \textbf{1.9} & 1.8 \\ \hline
Similarity ($\mathcal{T} = \mathcal{T}_{\text{sim}}$) & 0.6 & 1.1 & 1.5 & \textbf{1.6}\\ \hline
\end{tabular}
\end{center}
\caption{\label{tab:mnist_results} Accuracy and invariance scores of different classifers on the MNIST dataset.}
\end{table}
\end{small}
% END AFTER CORRECTION OF CODE

Table \ref{tab:mnist_results} reports the performance of the different classifiers under study, and their invariance scores $\hat{\rho}_{\mathcal{T}} (f)$ using Manitest. As expected, the linear and RBF-SVM classifiers compare poorly to other classifiers in terms of invariance. This is due to the construction of the CNN and Scat. PCA, which explicitly take into account the invariance through pooling operations, while others do not. Moreover, it can be noted that Scat. PCA outperforms CNN in terms of robustness to translations, and global similarity transformations, even if the two classifiers have similar test error. This result is in agreement with the theoretical evidence \cite{bruna2013invariant, Mallat2012} showing that scattering classifiers are invariant to deformations. 

\begin{figure}[t]
\centering
\includegraphics[width=0.9\textwidth]{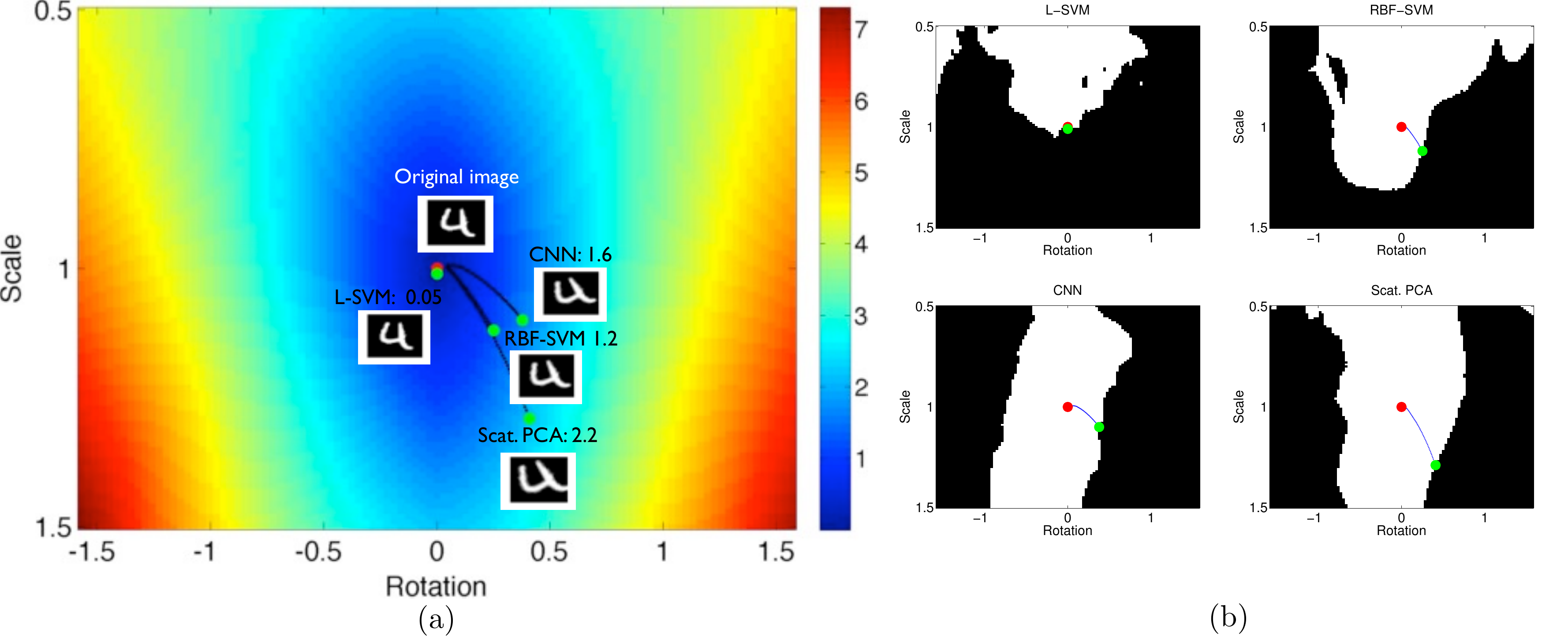}
\caption{\label{fig:distanceMap1} Distance map with $\mathcal{T}_{\text{dil+rot}}$ group (a), and correctly classified regions (b), for the four tested classifiers on an example image of digit ``4''. Geodesic paths are also shown.} % the color code indicates the geodesic distance from the identity transformation (shown by red dot at the center). For each classifier, the minimal transformation for which the output of the classifier is not correct (i.e., not ``4'') is indicated, along with the corresponding transformed image and geodesic path. \textit{b:} the region where the classifier correctly outputs the label ``4'' is shown in white. Geodesic paths are also shown.}
% Black indicates wrongly classifier, white indicates correctly classified. (Digit 4)}
\end{figure}

\begin{wrapfigure}{r}{0.4\textwidth}
  \begin{center}
    \includegraphics[width=0.4\textwidth]{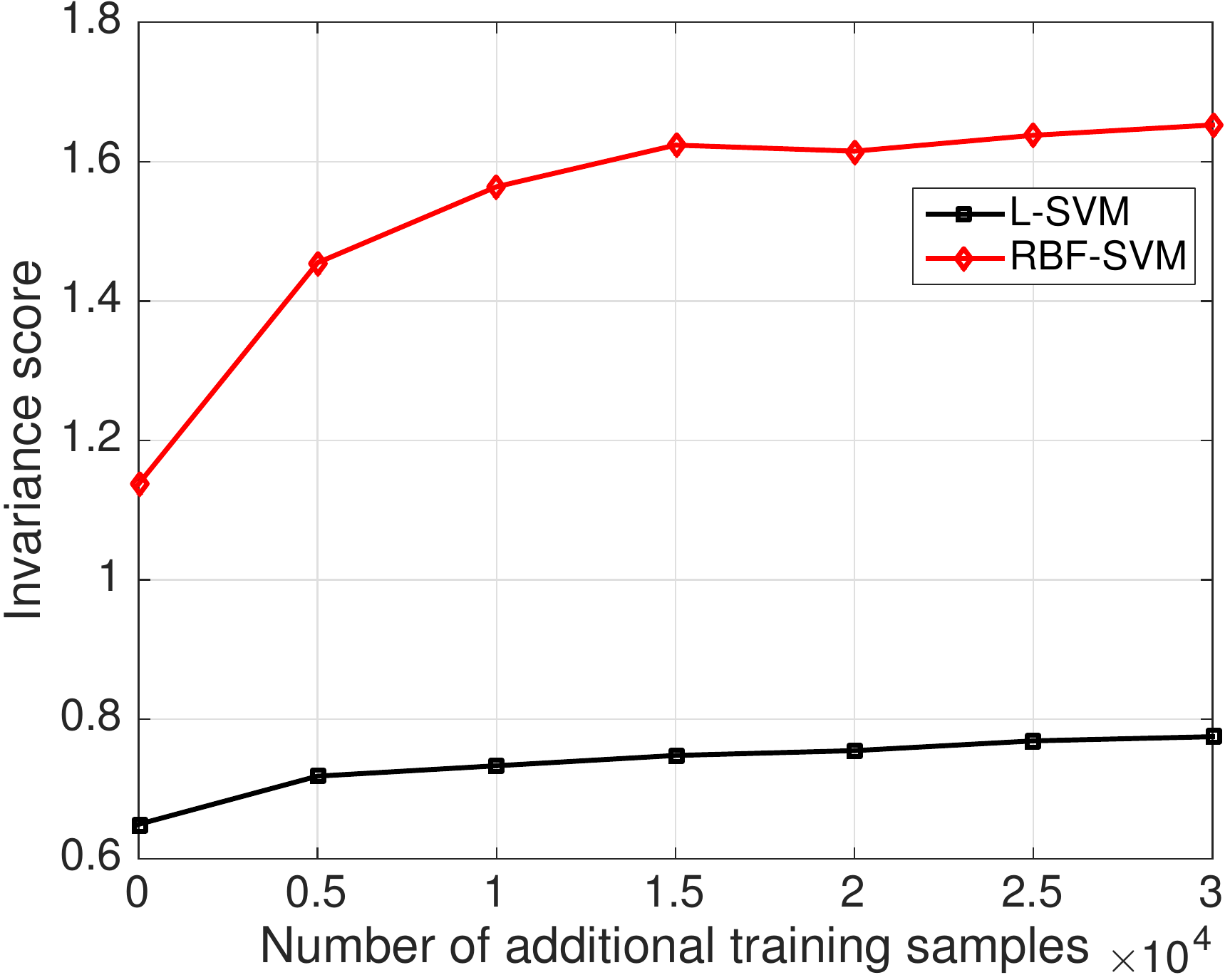}
  \end{center}
  \caption{\label{fig:influence_augmentation}Invariance score versus number of additional training samples, for MNIST, with $\mathcal{T} = \mathcal{T}_{\text{sim}}$.} % The additional training samples are generated by sampling at random a transformation in the similarity group with translation parameters of at most $3$ pixels in each direction, a scaling parameter between $0.7$, and $1.3$, and a rotation of at most $0.2$ radians.}
\end{wrapfigure}
To further get an insight on the invariance of the classifiers, we focus on the two-dimensional group $\mathcal{T}_{\text{dil+rot}}$, and show in Fig. \ref{fig:distanceMap1} (a) the geodesic distance map for an example image of digit ``4'' computed starting from the identity transformation (shown by a red dot at the center). Moreover, we overlay the minimally transformed images that change the labels of each of the classifiers, along with the corresponding geodesic paths. On this example, the Scat. PCA classifier is the most robust: a large dilation, accompanied with a rotation is required to change the classification label. In contrast, the linear SVM is easily ``fooled'' with a slight dilation. In Fig. \ref{fig:distanceMap1} (b) we illustrate in white the region of the Rotation-Scale plane, where the classifier outputs the correct label ``4''. 
Interestingly, the CNN and Scat. PCA classifiers are largely invariant to dilations (indicated by the vertical shape of the white region), while being moderately robust to rotations. % We refer to the supplementary material for more examples.

% MOVED TO SUPP MAT
%It should be noted that the geodesic distance map strongly depends on the considered image. To show this point, we do the same experiment as above, but using a ``0'' digit image in Fig. \ref{fig:distanceMap2}. Observe that the distance maps have very different shapes, which can be explained by the fact that the digit ``0'' only changes slightly as a rotation is applied, while the appearance of digit ``4'' is strongly affected by rotation.
% END MOVED TO SUPP MAT

% To show the dependence of the geodesic map on the considered image, we do the same experiment as above for a ``0'' digit image (Fig. \ref{fig:distanceMap2}). It should be noted that the distance maps have very different shapes, which can be explained by the fact that the digit ``0'' only changes slightly as a rotation is applied, while the appearance of digit ``4'' strongly changes with the rotation.

% RBF AUGMENTED
%\begin{figure}[t]
%\centering
%\includegraphics[width=0.3\textwidth]{dataAugmentationLinRBF.pdf}
%\caption{\label{fig:influence_augmentation}Invariance score $\hat{\rho}_{\mathcal{T}} (f)$ versus number of additional training samples, for MNIST, for $\mathcal{T} = \mathcal{T}_{\text{sim}}$. The additional training samples are generated by sampling at random a transfromation in the similarity group with translation parameters of at most $3$ pixels in each direction, a scaling parameter between $0.7$, and $1.3$, and a rotation of at most $0.2$ radians.}
%\end{figure}
% END RBF AUGMENTED

In vision tasks, it is common practice to augment the training data with artificial examples obtained by slightly distorting the original examples to achieve invariance. Although this practice is known to improve the classification performance of the classifiers on many tasks, its effect on the invariance of the classifier is not quantitatively understood. % We assess the gain in invariance using the proposed approach. 
Fig. \ref{fig:influence_augmentation} illustrates the Manitest invariance scores for L-SVM and RBF-SVM classifiers trained on augmented training sets obtained by randomly generating transformations\footnote{Random transformations are constrained as follows: translation of at most $3$ pixels in each direction, a scaling parameter between $0.7$, and $1.3$, and a rotation of at most $0.2$ radians.} from the similarity group $\mathcal{T}_{\text{sim}}$, on the MNIST dataset. Both classifiers improve their invariance score as more transformed samples are added to the training set. 
This result has moreover an element of surprise, as RBF-SVM succeeds in improving its invariance score by around $50\%$ with mere additions of artificial examples in the training set, and outperforms the invariance of CNN (without data augmentation). Moreover, the obtained score is comparable to Scat. PCA classifier, which is carefully designed to satisfy invariance properties.
This experiment permits to characterize the actual power of data augmentation for \textit{learning} the invariance from the data. 
\vspace{-4mm}
\subsection{Natural images}

In this second experimental section, we perform experiments on the CIFAR-10 dataset \cite{krizhevsky2009learning}. We focus on baseline CNN classifiers, and learn architectures with $1$, $2$ and $3$ hidden layers. Specifically, each layer consists of a successive combination of convolutional, rectified linear units and pooling operations. The convolutional layers consist of $5 \times 5$ filters with respectively $32, 32$ and $64$ feature maps for each layer, and the pooling operations are done on a window of size $3 \times 3$ with a stride parameter of $2$. We build the three architectures gradually, by successively stacking a new hidden layer on top of the previous architecture (kept fixed). The last hidden layer is then connected to a fully connected layer, and the softmax loss is used. Moreover, the different architectures are trained with stochastic gradient descent. On the test set, the error of the three architectures are respectively $35.6\%$, $25.0\%$ and $22.7\%$.

\begin{figure}[ht]
\centering
\subfigure[Translations]{
\includegraphics[width=0.3\textwidth]{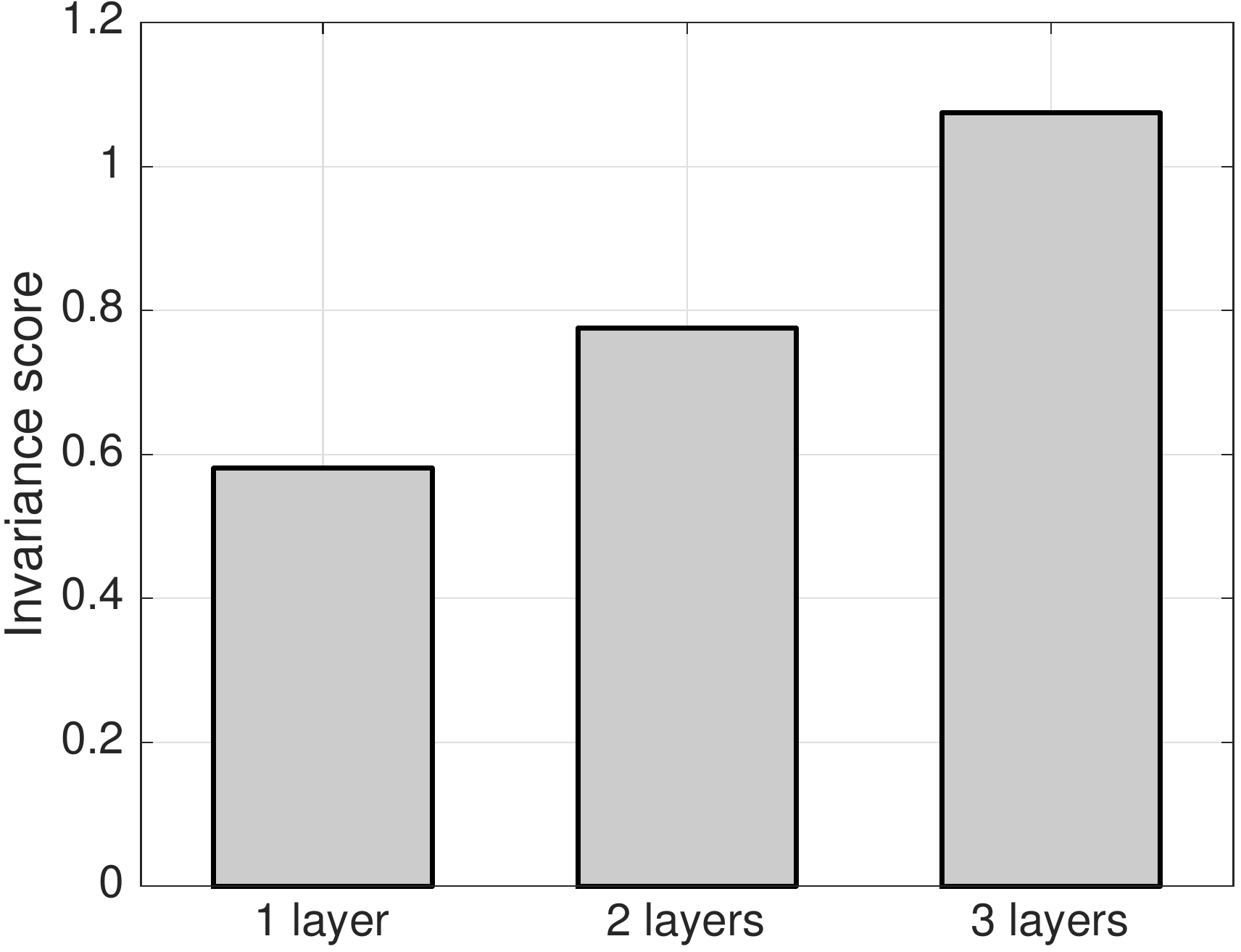}
}
\centering
\subfigure[Dilation + Rotation]{
\includegraphics[width=0.3\textwidth]{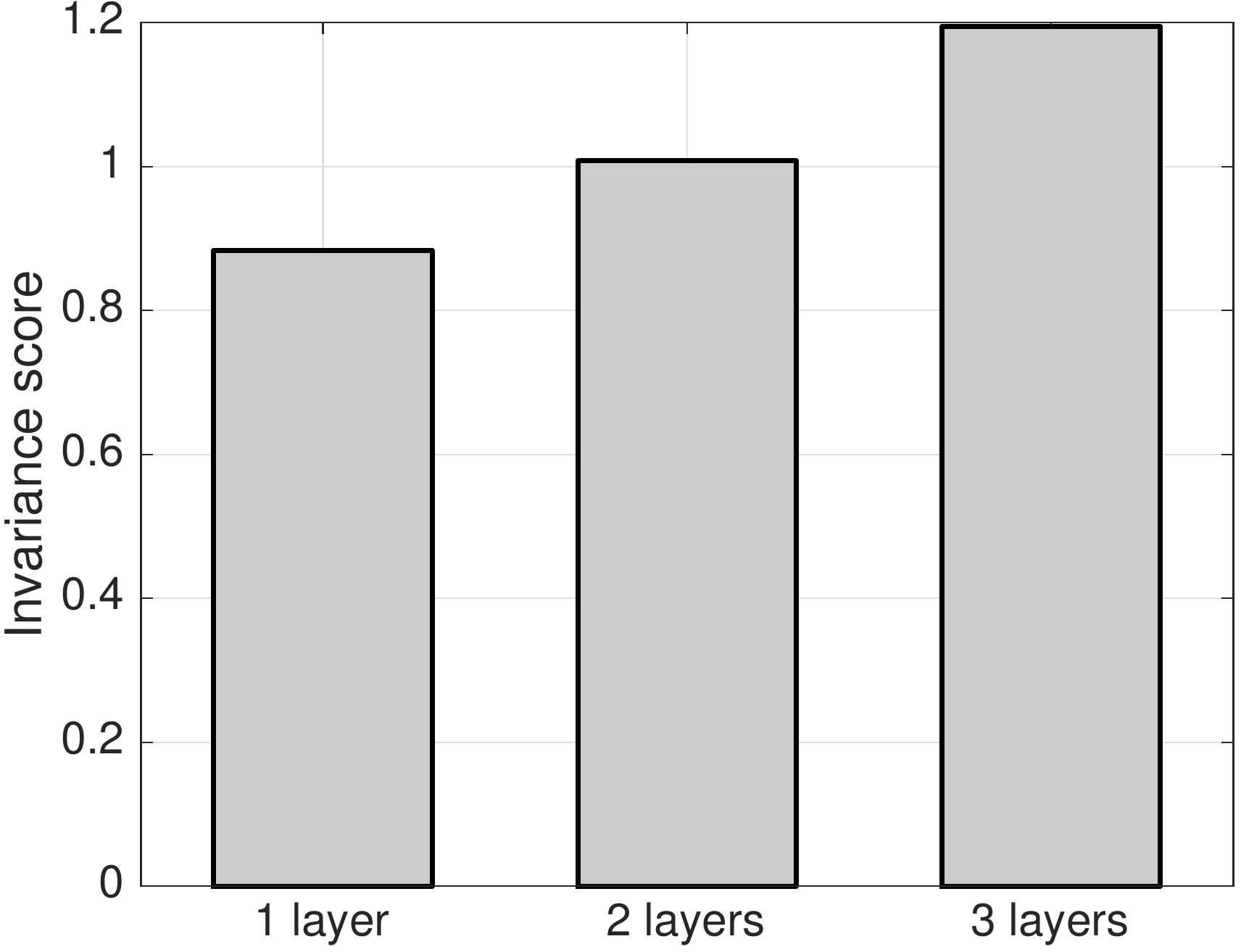}
}
\centering
\subfigure[Similarity]{
\includegraphics[width=0.3\textwidth]{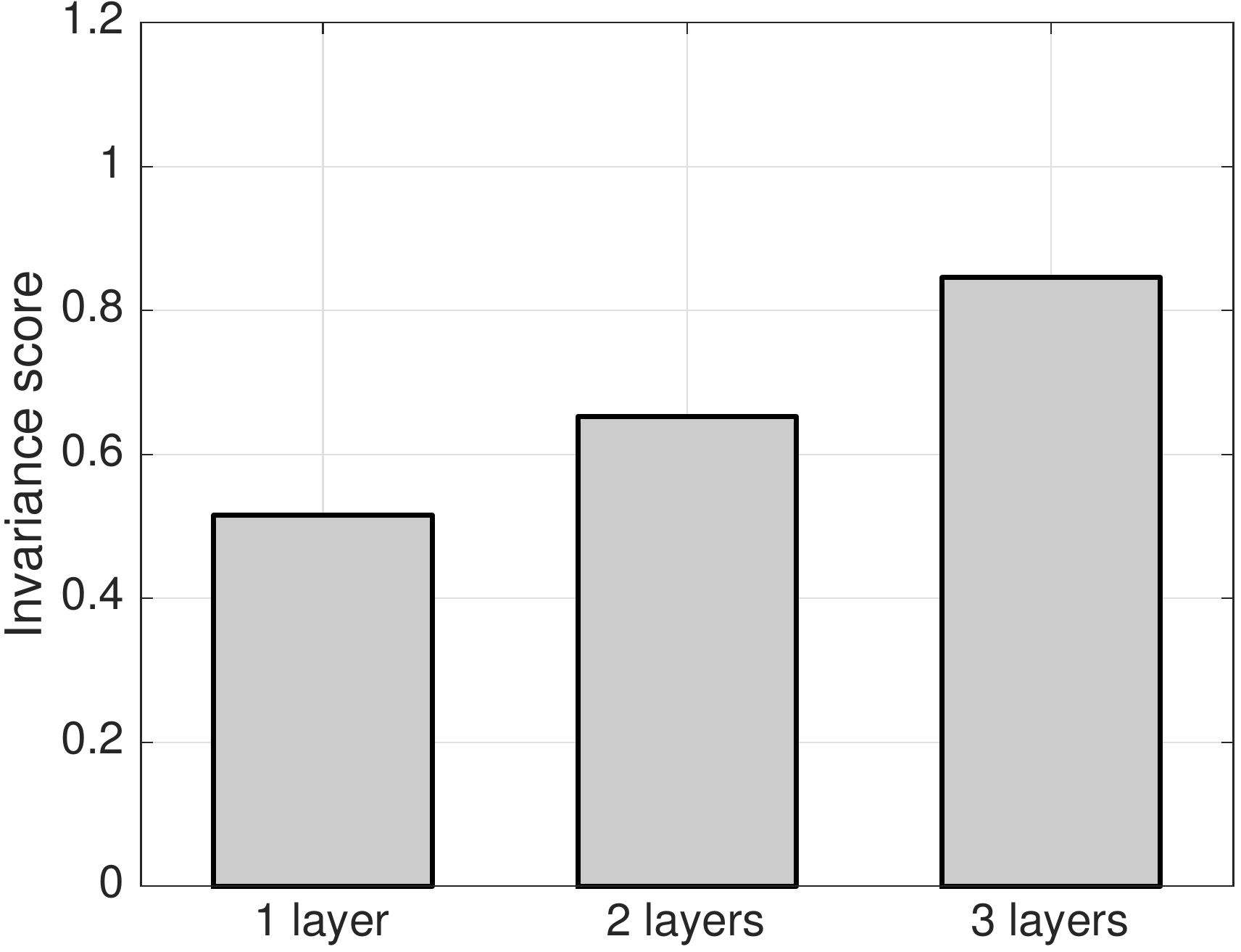}
}
\caption{\label{fig:hists_3layers}Invariance scores of CNNs on $\mathcal{T}_{\text{trans}}$, $\mathcal{T}_{\text{dil+rot}}$ and $\mathcal{T}_{\text{sim}}$, for the CIFAR-10 dataset.}
\end{figure}

%\begin{figure}[t]
%\centering
%\includegraphics[width=1\textwidth]{top_15_worst_15.pdf}
%\caption{\label{fig:top_worst_images}Top $15$ and worst $15$ images in terms of their invariances scores for $\mathcal{T} = \text{SIM} (2)$. The first rows show the original images, and the second rows the transformed images. The invariance score is indicated on the top left of each transformed image. All images are correctly classified by the 3-layer CNN. \red{Put in supplementary material on one page an evolution of all images, from worst to best}}
%\end{figure}

We show in Fig. \ref{fig:hists_3layers} the Manitest invariance scores of the three architectures. Our approach captures the \textit{increasing} invariance with the number of layers of the network, for the three groups under study. This result is in agreement with empirical studies and previous known belief \cite{goodfellow2009measuring, Bengio2012} that invariance increases with the depth of the network. However, while previous results were measuring the invariance with respect to a one dimensional transformation group (e.g., rotation only), Manitest provides a systematic and principled way of verifying the increased invariance of CNNs with depth on more complex Lie groups (e.g., similarity transformations).
% and Manitest provides a systematic and principled way of verifying this property. 
Interestingly enough, it should be noted that despite the relatively small difference in performance between the two and three layers architectures, the invariance score strongly increases. This highlights again that invariance and performance measures capture two different properties of classifiers.

%
%\begin{figure}[t]
%\centering
%\subfigure[Worst $3$]{
%\includegraphics[width=0.25\textwidth]{3leastRobust_images.pdf}
%}
%\centering
%\subfigure[Average $3$]{
%\includegraphics[width=0.25\textwidth]{3averageRobust_images.pdf}
%}
%\centering
%\subfigure[Top $3$]{
%\includegraphics[width=0.25\textwidth]{3mostRobust_images.pdf}
%}
%\caption{\label{fig:top_mean_worst} Illustration of images having (a) worst, (b) average, (c) top invariance to \textbf{similarity} transformations (i.e., $\mathcal{T} = \mathcal{T}_{\text{sim}}$), for the three-layer CNN. The first rows show the original images, and the second rows the minimally transformed images changing the prediction of the CNN. The Manitest invariance score is indicated on each transformed image. All original images are \textbf{correctly classified} by the 3-layer CNN. See supp. material for more images.}
%\end{figure}

% AFTER REVISION
\begin{figure}[t]
\centering
\subfigure[Worst $20$]{
\includegraphics[width=0.45\textwidth]{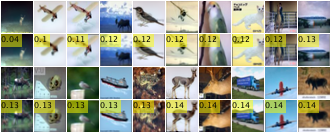}
}
\centering
\subfigure[Average $20$]{
\includegraphics[width=0.45\textwidth]{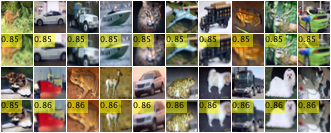}
}
\centering
\subfigure[Top $20$]{
\includegraphics[width=0.45\textwidth]{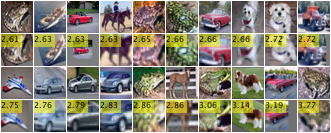}
}
\caption{\label{fig:top_mean_worst} Illustration of images having (a) worst, (b) average, (c) top invariance to \textbf{similarity} transformations (i.e., $\mathcal{T} = \mathcal{T}_{\text{sim}}$), for the three-layer CNN. The odd rows show the original images, and the even rows show the minimally transformed images changing the prediction of the CNN. The Manitest invariance score $\Delta_{\mathcal{T}} (I; f)$ is indicated on each transformed image. All original images are \textbf{correctly classified} by the 3-layer CNN.}
\end{figure}

% Robustness of worst, average and top images in terms of robustness in $\mathcal{T}_{\text{sim}}$.

%\begin{figure}[t]
%\centering
%\includegraphics[width=1\textwidth]{mostRobust_images.pdf}
%\caption{\label{fig:top_images}Top $45$ images in terms of their invariance scores for $\mathcal{T} = \text{SIM} (2)$, ranked in decreasing order. The odd rows show the original images, and the even rows the transformed images. The invariance score is indicated on the top left of each transformed image. All images are correctly classified by the 3-layer CNN.}
%\end{figure}
%
%\begin{figure}[t]
%\centering
%\includegraphics[width=1\textwidth]{leastRobust_images.pdf}
%\caption{\label{fig:worst_image}Worst $45$ images in terms of their invariance scores for $\mathcal{T} = \text{SIM} (2)$, ranked in increasing order. The odd rows show the original images, and the even rows the transformed images. The invariance score is indicated on the top left of each transformed image. All images are correctly classified by the 3-layer CNN.}
%\end{figure}

% \red{Define Invariance score above... Say that we are going to refer to both quantities (image-related, and not image-related) as invariance scores}

Compared to the handwritten digits task, note that the Manitest scores obtained on the CIFAR task are generally much smaller, which suggests that it is harder to achieve invariance on this task. To visualize the level of invariance of the 3-layer CNN on the CIFAR-10 dataset, we show in Fig. \ref{fig:top_mean_worst} sorted example images. For images with an average invariance score or less, note that the distinction between the transformed and original images are hardly perceptible. This suggests that the CNN is not robust to combinations of translations, rotation and dilation, even if it achieves a high accuracy. On the other hand, the difference between the original and the minimally transformed images are clearly perceptible for the top-scored images, even though a human observer is likely to correctly recognize the class of the transformed images. % We refer to supplementary material for more illustrations related to this experiment.

\vspace{-5mm}
\section{Conclusion}
\vspace{-2mm}
In this paper, we proposed a systematic and rigorous approach for measuring the invariance of any classifier to low-dimensional transformation groups. Using a manifold perspective, we were able to convert the problem of assessing the classifier's invariance to that of computing geodesic distances. Using Manitest, we quantified the increasing invariance of CNNs with depth, and highlighted the importance of data augmentation for learning invariance from data. We believe Manitest will be used to perform an in-depth empirical analysis of different classification architectures, in order to have a better understanding of the building blocks that best preserve invariance, and potentially build more robust classifiers.

%of the different components that constitute complex classifiers, (e.g., deep nets) in order to understand the invariance properties of the different building blocks.
%
%
%We believe the proposed generic approach for assessing the invariance of classifiers will lead to a better understanding of the classifiers that preserve invariance. For example, our tool can be used to compare 
%
%
%We presented a tool. Opens many possibilities for future research that will allow comparing different architectures (e.g., comparing max-pooling to average pooling, etc...) to reach better invariances, and lead us nearer to the human visual system (not only in terms of accuracy, but also robustness).

\vspace{2mm}
\begin{small}
\textbf{Acknowledgments.} We are grateful for the comments provided by the anonymous reviewers. We also thank Laurent Jacques and Gabriel Peyré for their insights on Fast Marching. We thank Luca Baroffio and Hamza Fawzi for their comments on the paper draft.
\end{small}

\newpage
\bibliography{refs}

\begin{thebibliography}{38}
\providecommand{\natexlab}[1]{#1}
\providecommand{\url}[1]{\texttt{#1}}
\expandafter\ifx\csname urlstyle\endcsname\relax
  \providecommand{\doi}[1]{doi: #1}\else
  \providecommand{\doi}{doi: \begingroup \urlstyle{rm}\Url}\fi

\bibitem[Bengio et~al.(2013)Bengio, Courville, and Vincent]{Bengio2012}
Yoshua Bengio, Aaron Courville, and Pascal Vincent.
\newblock Representation learning: A review and new perspectives.
\newblock \emph{IEEE Transactions on Pattern Analysis and Machine
  Intelligence}, 35\penalty0 (8):\penalty0 1798--1828, 2013.

\bibitem[Benmansour and Cohen(2011)]{benmansour2011tubular}
Fethallah Benmansour and Laurent~D Cohen.
\newblock Tubular structure segmentation based on minimal path method and
  anisotropic enhancement.
\newblock \emph{International Journal of Computer Vision}, 92\penalty0
  (2):\penalty0 192--210, 2011.

\bibitem[Berkes and Wiskott(2005)]{berkes2005slow}
Pietro Berkes and Laurenz Wiskott.
\newblock Slow feature analysis yields a rich repertoire of complex cell
  properties.
\newblock \emph{Journal of Vision}, 5\penalty0 (6):\penalty0 9, 2005.

\bibitem[Bruna(2012)]{Bruna2012}
Joan Bruna.
\newblock \emph{{Scattering representation for recognition}}.
\newblock PhD thesis, 2012.

\bibitem[Bruna and Mallat(2013)]{bruna2013invariant}
Joan Bruna and St{\'e}phane Mallat.
\newblock Invariant scattering convolution networks.
\newblock \emph{IEEE Transactions on Pattern Analysis and Machine
  Intelligence}, 35\penalty0 (8):\penalty0 1872--1886, 2013.

\bibitem[Chang and Lin(2011)]{chang2011libsvm}
Chih-Chung Chang and Chih-Jen Lin.
\newblock Libsvm: a library for support vector machines.
\newblock \emph{ACM Transactions on Intelligent Systems and Technology (TIST)},
  2\penalty0 (3):\penalty0 27, 2011.

\bibitem[Crandall and Lions(1983)]{crandall1983viscosity}
Michael~G Crandall and Pierre-Louis Lions.
\newblock Viscosity solutions of {H}amilton--{J}acobi equations.
\newblock \emph{Transactions of the American Mathematical Society},
  277\penalty0 (1):\penalty0 1--42, 1983.

\bibitem[Dijkstra(1959)]{dijkstra1959note}
Edsger~W Dijkstra.
\newblock A note on two problems in connexion with graphs.
\newblock \emph{Numerische mathematik}, 1\penalty0 (1):\penalty0 269--271,
  1959.

\bibitem[Donoho and Grimes(2005)]{donoho2005image}
David Donoho and Carrie Grimes.
\newblock Image manifolds which are isometric to euclidean space.
\newblock \emph{Journal of mathematical imaging and vision}, 23\penalty0
  (1):\penalty0 5--24, 2005.

\bibitem[Fan et~al.(2008)Fan, Chang, Hsieh, Wang, and Lin]{fan2008liblinear}
Rong-En Fan, Kai-Wei Chang, Cho-Jui Hsieh, Xiang-Rui Wang, and Chih-Jen Lin.
\newblock Liblinear: A library for large linear classification.
\newblock \emph{The Journal of Machine Learning Research}, 9:\penalty0
  1871--1874, 2008.

\bibitem[Fawzi and Frossard(2013)]{fawzi2013image}
Alhussein Fawzi and Pascal Frossard.
\newblock Image registration with sparse approximations in parametric
  dictionaries.
\newblock \emph{SIAM Journal on Imaging Sciences}, 6\penalty0 (4):\penalty0
  2370--2403, 2013.

\bibitem[Geusebroek et~al.(2005)Geusebroek, Burghouts, and
  Smeulders]{geusebroek2005amsterdam}
Jan-Mark Geusebroek, Gertjan~J Burghouts, and Arnold~WM Smeulders.
\newblock The {A}msterdam library of object images.
\newblock \emph{International Journal of Computer Vision}, 61\penalty0
  (1):\penalty0 103--112, 2005.

\bibitem[Goodfellow et~al.(2009)Goodfellow, Lee, Le, Saxe, and
  Ng]{goodfellow2009measuring}
Ian Goodfellow, Honglak Lee, Quoc~V Le, Andrew Saxe, and Andrew~Y Ng.
\newblock Measuring invariances in deep networks.
\newblock In \emph{Advances in neural information processing systems}, pages
  646--654, 2009.

\bibitem[Jacques and De~Vleeschouwer(2008)]{jacques2008geometrical}
Laurent Jacques and Christophe De~Vleeschouwer.
\newblock A geometrical study of matching pursuit parametrization.
\newblock \emph{IEEE Transactions on Signal Processing}, 56\penalty0
  (7):\penalty0 2835--2848, 2008.

\bibitem[Kimmel and Sethian(1998)]{kimmel1998computing}
Ron Kimmel and James~A Sethian.
\newblock Computing geodesic paths on manifolds.
\newblock \emph{Proceedings of the National Academy of Sciences}, 95\penalty0
  (15):\penalty0 8431--8435, 1998.

\bibitem[Kokiopoulou and Frossard(2009)]{kokiopoulou2009minimum}
Effrosyni Kokiopoulou and Pascal Frossard.
\newblock Minimum distance between pattern transformation manifolds: Algorithm
  and applications.
\newblock \emph{IEEE Transactions on Pattern Analysis and Machine
  Intelligence}, 31\penalty0 (7):\penalty0 1225--1238, 2009.

\bibitem[Krizhevsky and Hinton(2009)]{krizhevsky2009learning}
Alex Krizhevsky and Geoffrey Hinton.
\newblock Learning multiple layers of features from tiny images.
\newblock \emph{Master's thesis, Department of Computer Science, University of
  Toronto}, 2009.

\bibitem[Krizhevsky et~al.(2012)Krizhevsky, Sutskever, and
  Hinton]{krizhevsky2012imagenet}
Alex Krizhevsky, Ilya Sutskever, and Geoffrey~E Hinton.
\newblock Imagenet classification with deep convolutional neural networks.
\newblock In \emph{Advances in neural information processing systems}, pages
  1097--1105, 2012.

\bibitem[Larochelle et~al.(2009)Larochelle, Bengio, Louradour, and
  Lamblin]{larochelle2009exploring}
Hugo Larochelle, Yoshua Bengio, J{\'e}r{\^o}me Louradour, and Pascal Lamblin.
\newblock Exploring strategies for training deep neural networks.
\newblock \emph{The Journal of Machine Learning Research}, 10:\penalty0 1--40,
  2009.

\bibitem[LeCun et~al.(1998)LeCun, Bottou, Bengio, and
  Haffner]{lecun1998gradient}
Yann LeCun, Leon Bottou, Yoshua Bengio, and Patrick Haffner.
\newblock Gradient-based learning applied to document recognition.
\newblock \emph{\textit{Proceedings of the IEEE}}, 86\penalty0 (11):\penalty0
  2278--2324, 1998.

\bibitem[LeCun et~al.(2004)LeCun, Huang, and Bottou]{lecun2004learning}
Yann LeCun, Fu~Jie Huang, and Leon Bottou.
\newblock Learning methods for generic object recognition with invariance to
  pose and lighting.
\newblock In \emph{IEEE Conference on Computer Vision and Pattern Recognition},
  volume~2, pages 97--104, 2004.

\bibitem[Lenc and Vedaldi(2015)]{lenc15understanding}
Karel Lenc and Andrea Vedaldi.
\newblock Understanding image representations by measuring their equivariance
  and equivalence.
\newblock In \emph{{IEEE} Conference on Computer Vision and Pattern Recognition
  ({CVPR})}, 2015.

\bibitem[Lin(2003)]{lin2003enhancement}
Qingfen Lin.
\newblock Enhancement, extraction, and visualization of 3d volume data.
\newblock \emph{PhD thesis}, 2003.

\bibitem[Mallat(2012)]{Mallat2012}
St{\'e}phane Mallat.
\newblock Group invariant scattering.
\newblock \emph{Communications on Pure and Applied Mathematics}, 65\penalty0
  (10):\penalty0 1331--1398, 2012.

\bibitem[Mirebeau(2014)]{mirebeau2014anisotropic}
Jean-Marie Mirebeau.
\newblock Anisotropic fast-marching on cartesian grids using lattice basis
  reduction.
\newblock \emph{SIAM Journal on Numerical Analysis}, 52\penalty0 (4):\penalty0
  1573--1599, 2014.

\bibitem[Peyr{\'e} et~al.(2010)Peyr{\'e}, P{\'e}chaud, Keriven, and
  Cohen]{peyre2010geodesic}
Gabriel Peyr{\'e}, Micka{\"e}l P{\'e}chaud, Renaud Keriven, and Laurent~D
  Cohen.
\newblock Geodesic methods in computer vision and graphics.
\newblock \emph{Foundations and Trends in Computer Graphics and Vision},
  5\penalty0 (3--4):\penalty0 197--397, 2010.

\bibitem[Sethian and Vladimirsky(2000)]{sethian2000fast}
James~A Sethian and Alexander Vladimirsky.
\newblock Fast methods for the eikonal and related {H}amilton--{J}acobi
  equations on unstructured meshes.
\newblock \emph{Proceedings of the National Academy of Sciences}, 97\penalty0
  (11):\penalty0 5699--5703, 2000.

\bibitem[Sethian and Vladimirsky(2003)]{sethian2003ordered}
James~A Sethian and Alexander Vladimirsky.
\newblock Ordered upwind methods for static {H}amilton--{J}acobi equations:
  Theory and algorithms.
\newblock \emph{SIAM Journal on Numerical Analysis}, 41\penalty0 (1):\penalty0
  325--363, 2003.

\bibitem[Simard et~al.(1998)Simard, LeCun, Denker, and
  Victorri]{simard1998transformation}
Patrice~Y Simard, Yann~A LeCun, John~S Denker, and Bernard Victorri.
\newblock Transformation invariance in pattern recognition--tangent distance
  and tangent propagation.
\newblock In \emph{Neural networks: tricks of the trade}, pages 239--274.
  Springer, 1998.

\bibitem[Soatto and Chiuso(2015)]{soattoC15}
Stefano Soatto and Alessandro Chiuso.
\newblock Visual scene representations: Sufficiency, minimality, invariance and
  deep approximation.
\newblock In \emph{International Conference on Learning Representations (ICLR)
  Workshop}. 2015.

\bibitem[Sohn and Lee(2012)]{sohn2012learning}
Kihyuk Sohn and Honglak Lee.
\newblock Learning invariant representations with local transformations.
\newblock In \emph{International Conference on Machine Learning}, 2012.

\bibitem[Szegedy et~al.(2014)Szegedy, Zaremba, Sutskever, Bruna, Erhan,
  Goodfellow, and Fergus]{szegedy2013intriguing}
Christian Szegedy, Wojciech Zaremba, Ilya Sutskever, Joan Bruna, Dumitru Erhan,
  Ian Goodfellow, and Rob Fergus.
\newblock Intriguing properties of neural networks.
\newblock In \emph{International Conference on Learning Representations}, 2014.

\bibitem[Taigman et~al.(2014)Taigman, Yang, Ranzato, and
  Wolf]{taigman2014deepface}
Yaniv Taigman, Ming Yang, Marc'Aurelio Ranzato, and Lior Wolf.
\newblock Deepface: Closing the gap to human-level performance in face
  verification.
\newblock In \emph{IEEE Conference on Computer Vision and Pattern Recognition},
  pages 1701--1708, 2014.

\bibitem[Tenenbaum et~al.(2000)Tenenbaum, De~Silva, and
  Langford]{tenenbaum2000global}
Joshua~B Tenenbaum, Vin De~Silva, and John~C Langford.
\newblock A global geometric framework for nonlinear dimensionality reduction.
\newblock \emph{Science}, 290\penalty0 (5500):\penalty0 2319--2323, 2000.

\bibitem[Tsitsiklis(1995)]{tsitsiklis1995efficient}
John~N Tsitsiklis.
\newblock Efficient algorithms for globally optimal trajectories.
\newblock \emph{IEEE Transactions on Automatic Control}, 40\penalty0
  (9):\penalty0 1528--1538, 1995.

\bibitem[Vasconcelos and Lippman(1998)]{vasconcelos1998multiresolution}
Nuno Vasconcelos and Andrew Lippman.
\newblock Multiresolution tangent distance for affine-invariant classification.
\newblock \emph{Advances in neural information processing systems}, pages
  843--849, 1998.

\bibitem[Vedaldi and Lenc(2014)]{vedaldi2014matconvnet}
Andrea Vedaldi and Karel Lenc.
\newblock Matconvnet-convolutional neural networks for matlab.
\newblock \emph{arXiv preprint arXiv:1412.4564}, 2014.

\bibitem[Wakin et~al.(2005)Wakin, Donoho, Choi, and
  Baraniuk]{wakin2005multiscale}
Michael~B Wakin, David~L Donoho, Hyeokho Choi, and Richard~G Baraniuk.
\newblock The multiscale structure of non-differentiable image manifolds.
\newblock In \emph{Wavelets XI in SPIE International Symposium on Optical
  Science and Technology}, 2005.

\end{thebibliography}
\end{document}